\documentclass[letterpaper, 10 pt, conference]{ieeeconf}

\IEEEoverridecommandlockouts

\usepackage{graphics}
\usepackage{epsfig}
\usepackage{times}
\usepackage{amsmath}
\usepackage{amssymb}
\usepackage{balance}
\usepackage[hidelinks]{hyperref}
\usepackage[table]{xcolor}
\usepackage{booktabs, colortbl, siunitx}
\usepackage[colorinlistoftodos]{todonotes}
\usepackage{booktabs}
\usepackage{multirow}

\definecolor{bestcolor}{HTML}{D9EAD3}
\definecolor{secondcolor}{HTML}{FFF2CC}

\newcommand{\best}{\cellcolor{bestcolor}}
\newcommand{\snd}{\cellcolor{secondcolor}}

\usepackage[ruled,vlined,linesnumbered]{algorithm2e}
\SetKwInput{KwInput}{Input}
\SetKwInput{KwOutput}{Output}
\usepackage{etoolbox}

\AtBeginEnvironment{thebibliography}{%
  \setlength{\itemsep}{0pt}%
  \setlength{\parskip}{0pt}%
}

\appto\normalsize{%
  \setlength{\abovedisplayskip}{3pt plus 1pt minus 1pt}%
  \setlength{\belowdisplayskip}{3pt plus 1pt minus 1pt}%
  \setlength{\abovedisplayshortskip}{0pt plus 1pt}%
  \setlength{\belowdisplayshortskip}{1.5pt plus 1pt minus 1pt}%
}

\makeatletter
\patchcmd{\@makecaption}
  {\scshape}
  {}{}{} 
\makeatother

\makeatletter
\patchcmd{\@makecaption}
  {\\}
  {. \ }{}{}
\makeatother

\title{\LARGE \bf
FGGS-LiDAR: Ultra-Fast, GPU-Accelerated Simulation from General 3DGS Models to LiDAR
}

\author{Junzhe Wu$^{1*}$, Yufei Jia$^{2*}$, Yiyi Yan$^{3}$, Zhixing Chen$^{1}$, Tiao Tan$^{1}$, Zifan Wang$^{4}$, Guangyu Wang$^{1}$, \\
BoKui Chen$^{1\dag}$, Guyue Zhou$^{5\dag}$
\thanks{*Equal contribution; \dag Corresponding Authors.}
\thanks{$^{1}$Tsinghua University, Shenzhen, China.
        {\tt\small \{wjz25, chenzx24, tt23, wanggy24\}@mails.tsinghua.edu.cn, chenbk@tsinghua.edu.cn}}%
\thanks{$^{2}$Department of Electronic Engineering, Tsinghua University, Beijing, China.
        {\tt\small jyf23@mails.tsinghua.edu.cn}}%
\thanks{$^{3}$DISCOVER Robotics.
        {\tt\small albertyanyy@gmail.com}}
\thanks{$^{4}$Hong Kong University of Science and Technology (Guangzhou).
        {\tt\small wang\_zifan@outlook.com}}%
\thanks{$^{5}$The Institute for AI Industry Research (AIR), Tsinghua University, Beijing, China.
        {\tt\small zhouguyue@air.tsinghua.edu.cn}}%
}

\usepackage{capt-of}

\let\oldtwocolumn\twocolumn
\renewcommand\twocolumn[1][]{%
  \oldtwocolumn[{#1}{
    \begin{center}
      \includegraphics[width=\textwidth]{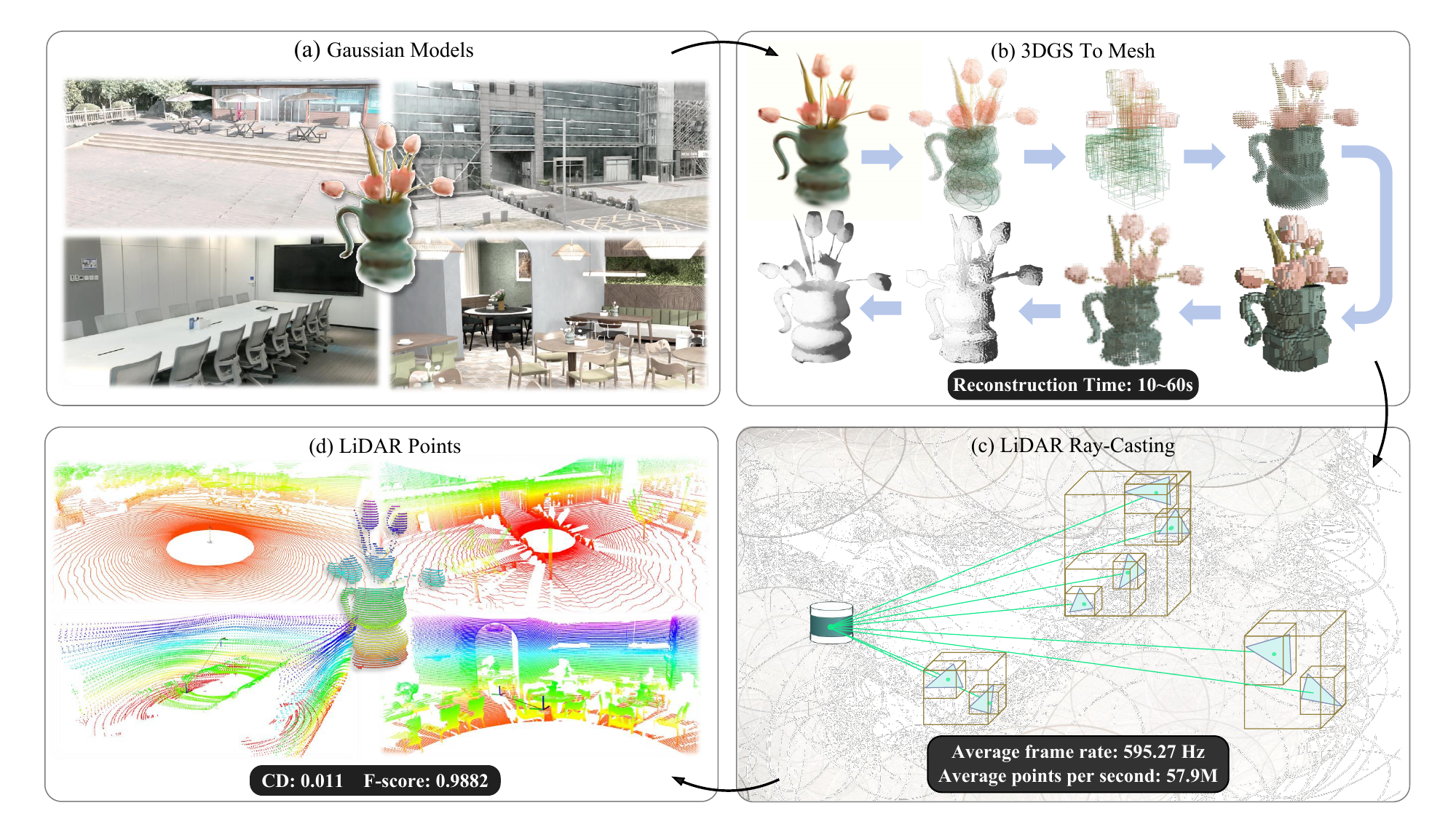}
      \captionof{figure}{\textbf{Overview of FGGS-LiDAR framework.} 
  (a) Visualization of a generic 3DGS dataset. 
  (b) Schematic illustration of our 3DGS2Mesh conversion pipeline, which transforms Gaussian primitives into mesh-based scene representations. (We use the vase as an example for the illustration.)
  (c) Conceptual diagram of our LiDAR ray-casting procedure, simulating range measurements through efficient GPU-based traversal.
  (d) Visualization of rendered LiDAR scans.}
      \label{fig:teaser}
    \end{center}
  }]
}

\usepackage{caption}
\begin{document}

\maketitle

\thispagestyle{empty}
\pagestyle{empty}

\begin{abstract}
While 3D Gaussian Splatting (3DGS) has emerged as a strong representation for photorealistic rendering, its vast ecosystem of assets remains difficult to use for high-performance LiDAR simulation, a critical tool for robotics and autonomous driving. We present \textbf{FGGS-LiDAR}, a geometry-first framework that bridges this gap in a plug-and-play manner. Our method converts pretrained 3DGS assets into watertight meshes directly from Gaussian parameters, without requiring LiDAR-specific supervision or architectural alterations, via volumetric discretization and Truncated Signed Distance Field (TSDF) extraction. We pair this with a GPU-accelerated ray-casting module that simulates LiDAR returns at over 500 FPS and supports batched multi-environment simulation with up to 4096 environments. In large-scale parallel settings, FGGS-LiDAR achieves an order-of-magnitude lower LiDAR simulation latency than Isaac Sim. We validate FGGS-LiDAR on both indoor and outdoor scenes, demonstrating high LiDAR-simulation fidelity. Furthermore, on COLMAP-posed indoor benchmarks, we compare against existing 3DGS-to-mesh baselines and report lower LiDAR-simulation error. Code is at \url{https://github.com/discoverse-dev/FGGS-LiDAR}.
\end{abstract}

\section{INTRODUCTION}

LiDAR is a cornerstone modality for 3D perception, underpinning autonomous driving, localization, odometry, mapping, and indoor navigation~\cite{zhang2024lidar,yin2024survey,xu2022fast,charroud2024localization,savva2017minos}. To mitigate the prohibitive expense and logistical challenges of curating large-scale real-world datasets, simulation offers a controllable and reproducible source of data for training and benchmarking perception algorithms. Consequently, the fidelity, efficiency, and scalability of LiDAR simulators are critical factors that directly determine their utility in both research and real-world deployment.

The landscape of LiDAR simulation has historically been defined by a trade-off between asset creation and rendering performance. Traditional simulation pipelines, built on explicit mesh representations, can produce geometrically accurate and controllable data but are fundamentally bottlenecked by the need for high-quality, often manually created, 3D assets~\cite{manivasagam2020lidarsim,shah2017airsim,li2022pcgen,dosovitskiy2017carla}. This dependency on specialized assets limits their scalability and adaptability to diverse, real-world environments. In response, recent advances in neural fields, particularly Neural Radiance Fields (NeRF), have enabled the reconstruction of scenes directly from sensor data~\cite{mildenhall2021nerf}. However, while NeRF-based LiDAR simulators achieve impressive fidelity, their reliance on implicit representations and exhaustive volumetric ray marching renders them computationally intensive, with slow training times and low inference throughput that preclude real-time applications~\cite{zheng2024lidar4d,yang2023unisim,xue2024geonlf,zhang2024nerf,huang2023neural}.

The emergence of 3D Gaussian Splatting (3DGS) promised to resolve this impasse, offering a representation that combines photorealistic quality with real-time rendering speeds~\cite{kerbl20233dgs,liu2024citygaussian,feng2025flashgs,InteriorGS2025}. Yet, the standard 3DGS formulation, optimized for visual fidelity, is ill-suited for geometric sensing tasks: its rendering process often produces blurred surfaces and incoherent depth estimates, failing to model precise LiDAR first returns~\cite{chen2024pgsr,radl2024stopthepop,Huang2DGS2024,qian20243dgsavatar,yu2024mip}. Specialized Gaussian-based LiDAR methods~\cite{zhou2024lidarrt,jiang2025gslidar} address this by learning LiDAR-specific attributes, but typically require LiDAR-supervised retraining and often modify the standard 3DGS formulation. As a result, they are not directly applicable to arbitrary off-the-shelf, photometrically trained 3DGS assets, leaving the problem of plug-and-play LiDAR simulation from pretrained 3DGS assets largely open.

To bridge this gap, we introduce FGGS-LiDAR, a geometry-first framework that makes arbitrary pretrained 3DGS assets directly usable for LiDAR simulation. Unlike prior 3DGS-to-mesh pipelines that typically follow a pose-dependent multi-view depth/occupancy rendering $\rightarrow$ volumetric fusion route, whose reconstruction quality and computational cost depend on view coverage, depth consistency, and pose accuracy, our method recovers geometry directly from Gaussian parameters, i.e., means, covariances, and opacities, without view sampling, rendered depth maps, or external camera poses. We then reconstruct a narrow-band TSDF, extract a watertight mesh, and perform GPU ray-casting for efficient LiDAR simulation. Rather than claiming a new meshing primitive, TSDF, or ray-tracing algorithm, we present a geometry-first system that recovers geometry directly from Gaussian parameters, rather than rendered views, and couples it with high-throughput GPU simulation. For fair comparison with prior pose-dependent baselines, we additionally evaluate on COLMAP-posed benchmarks.

Finally, we contribute a highly optimized GPU ray-casting module that simulates LiDAR returns at over 500 FPS while supporting massively parallel batched simulation across thousands of environments on a single GPU.

\noindent\textbf{Our contributions are summarized as follows:}
\begin{itemize}
    \item \textbf{Geometry-first LiDAR simulation from 3DGS.}
    We propose a pose-free geometry recovery pipeline that converts pretrained 3DGS models into watertight meshes directly from Gaussian parameters, avoiding the commonly used multi-view depth/occupancy rendering and fusion route. This enables LiDAR simulation from arbitrary pretrained 3DGS assets without requiring COLMAP poses, rendered depth maps, or LiDAR supervision.

    \item \textbf{Efficient Gaussian-to-geometry conversion.}
    We introduce a GPU-accelerated voxelization framework using Gaussian AABBs and Morton-sorted LBVH spatial indexing, followed by narrow-band TSDF reconstruction that produces topology-consistent watertight meshes while maintaining scalability for large scenes.

    \item \textbf{Massively parallel LiDAR simulation.}
    We implement a plug-and-play, GPU-based batched ray-casting engine that supports thousands of environments (up to 4096 environments) in parallel and achieves ultra-fast LiDAR simulation (\textgreater 500 FPS), substantially outperforming general-purpose simulators such as Isaac Sim.

\end{itemize}

\section{Related Work}
We build upon 3D Gaussian Splatting (3DGS)~\cite{kerbl20233dgs} as a pretrained scene representation and focus on geometry conversion and LiDAR simulation.

\subsection{Mesh reconstruction from 3DGS}
Mesh reconstruction from 3DGS extends Gaussian representations to geometry-oriented applications. Representative pipelines such as GS2Mesh~\cite{wolf2024gs2mesh} and MILo~\cite{guavsdon2025milo} reconstruct meshes from Gaussian primitives via volumetric sampling, implicit surface extraction, and standard meshing. However, these methods often rely on auxiliary multiview reconstructions (e.g., COLMAP~\cite{schoenberger2016sfm,schoenberger2016mvs}) to provide camera poses and depth priors, coupling mesh extraction to the original image acquisition and training process. This dependency limits direct application to arbitrary pretrained 3DGS assets where such priors may be unavailable or inconsistent.

\subsection{Gaussian-based LiDAR simulation}
Gaussian representations have also been explored for LiDAR synthesis. LiDAR-RT~\cite{zhou2024lidarrt} performs Gaussian-based ray tracing with learnable LiDAR attributes (e.g., intensity and ray-drop probability), while GS-LiDAR~\cite{jiang2025gslidar} models LiDAR novel-view synthesis via panoramic Gaussian projection and explicit ray--Gaussian interactions. Despite promising results, existing Gaussian-based LiDAR approaches typically require supervision from real LiDAR scans and introduce task-specific extensions to standard 3DGS, reducing drop-in compatibility with large pretrained 3DGS models. Moreover, ray-traced formulations may incur higher computational cost, whereas rasterization-based formulations may not explicitly model time-of-flight. These gaps motivate a general and efficient LiDAR simulation pipeline that remains fully compatible with pretrained 3DGS and scales across domains.

\begin{figure*}[!t]
\vspace{4pt}
  \centering
  \includegraphics[width=\textwidth]{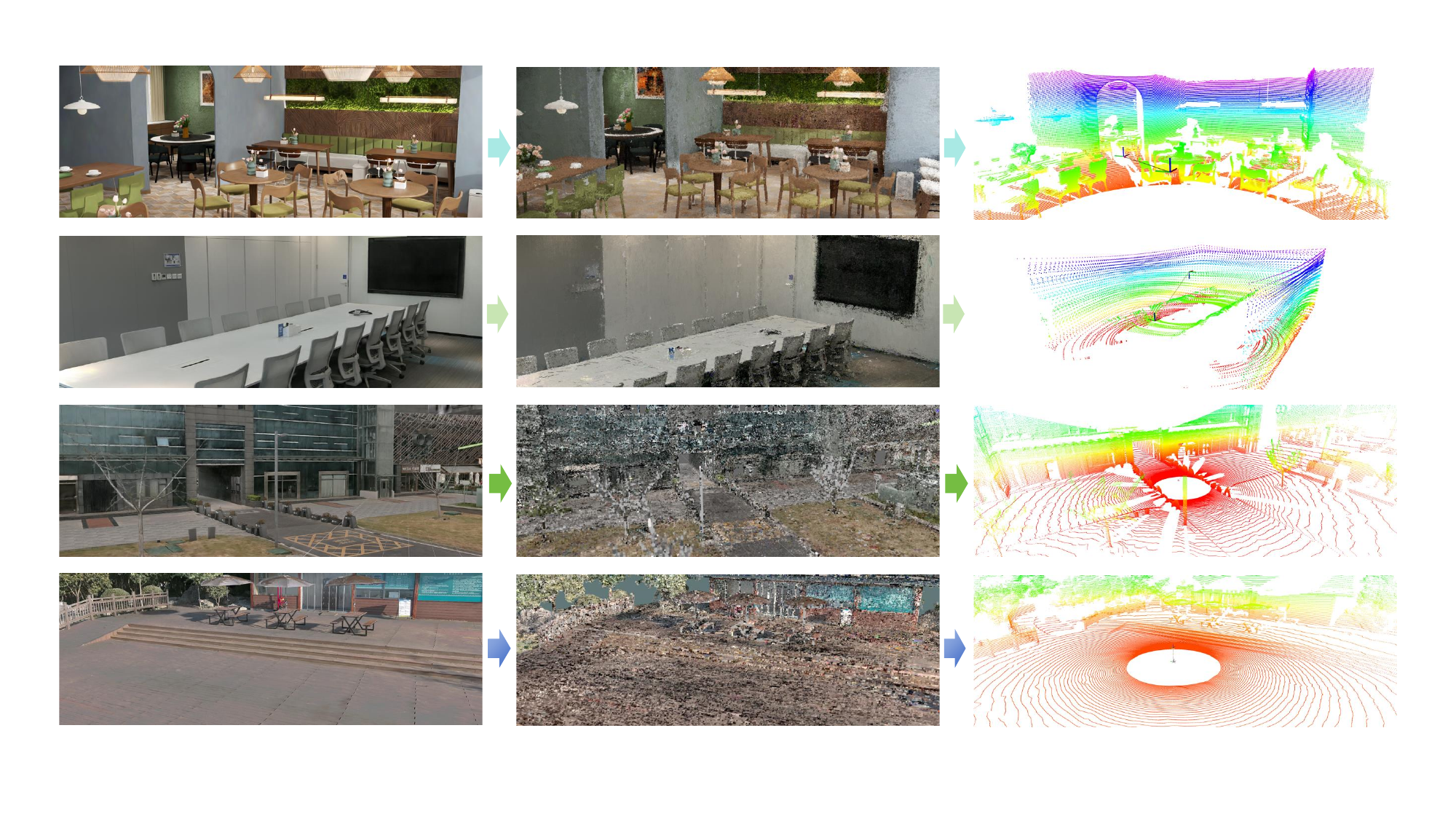}
    \caption{\textbf{LiDAR visualization across scenes.}
      A $4\times 3$ panel: each row corresponds to one scene. From left to right, the columns show (1) 3DGS visualization, (2) the voxelized mesh, and (3) the LiDAR point cloud rendered by FGGS-LiDAR.}
    \label{fig:rendering}
\end{figure*}

\section{Preliminaries}

\subsection{3D Gaussian Splatting (3DGS)}
We denote a pretrained 3DGS scene as a set of anisotropic Gaussians $\{G_i\}$~\cite{kerbl20233dgs}.
Each $G_i$ is parameterized by a mean $\mu_i\!\in\!\mathbb{R}^3$, a covariance $\Sigma_i\!\in\!\mathbb{R}^{3\times 3}$, and an opacity $\alpha_i\!\in\!(0,1)$.
Its spatial weight at $x\in\mathbb{R}^3$ is
\begin{equation}
w_i(x)=\exp\!\Big(-\tfrac{1}{2}(x-\mu_i)^\top \Sigma_i^{-1}(x-\mu_i)\Big),
\end{equation}
where $\Sigma_i = R_i S_i S_i^\top R_i^\top$ with a rotation $R_i$ and a (diagonal) scaling matrix $S_i$.
We only use the geometric parameters $(\mu_i,\Sigma_i,\alpha_i)$ in our pipeline.

\subsection{Truncated Signed Distance Function (TSDF)}
Given a surface $\partial\Omega$, the signed distance at $x$ is $d(x)=\operatorname{sgn}(x)\min_{y\in\partial\Omega}\|x-y\|$.
A TSDF truncates it with band radius $r$~\cite{curless1996volumetric}:
\begin{equation}
\phi(x)=\mathrm{clip}\big(d(x),-r,r\big),
\end{equation}
whose zero-level set $\{x\mid \phi(x)=0\}$ defines the surface for watertight mesh extraction.

\section{Method}

Our goal is to make a pretrained 3DGS asset directly usable for LiDAR simulation by converting it into a geometry representation that supports stable surface recovery and efficient first-hit queries. Standard 3DGS is a continuous and unstructured rendering representation: it is effective for photometric synthesis, but it does not directly provide a robust interface for watertight surface extraction or repeated LiDAR ray intersection. Rather than introducing a new meshing primitive in isolation, the contribution here is an integrated geometry-first system that couples direct Gaussian-parameter geometry recovery with high-throughput GPU simulation.

FGGS-LiDAR resolves this representation mismatch through three linked representation transitions. Sec.~\ref{sec1} converts Gaussian parameters into a sparse geometric proxy using LBVH-accelerated voxel queries, exposing where geometric support exists while avoiding global Gaussian--voxel evaluation. Sec.~\ref{sec2} turns this proxy into a watertight, simulation-ready surface through narrow-band TSDF reconstruction and mesh optimization. Sec.~\ref{sec_lidar} then realizes efficient LiDAR simulation through hierarchy-accelerated GPU ray-casting.

\subsection{LBVH-Accelerated Gaussian-to-Occupancy Conversion} \label{sec1}

The reconstruction system requires an explicit geometric support for surface recovery, whereas the input 3DGS asset is a continuous and unstructured set of Gaussian primitives. Direct evaluation of all Gaussians on a dense voxel grid is computationally prohibitive, with complexity $O(|\mathcal{V}|\,N)$. We therefore build a GPU linear bounding volume hierarchy (LBVH) over Gaussian AABBs and use tile-level culling to localize Gaussian--voxel evaluation.

Concretely, we partition the scene domain $\Omega \subset \mathbb{R}^3$ into an isotropic voxel grid with spacing $h$ and voxel centers $v_{ijk}$. For each $v_{ijk}$, we accumulate density $D(v_{ijk})$ from nearby Gaussians (weighted by squared opacity $\alpha_i^2$ to emphasize stable geometric support) and threshold it with a density threshold $\theta$ to obtain a raw occupancy grid $V_{\text{occ}}$. We then retain only surface voxels via $\mathrm{Surf}(i,j,k)$ and define the cleaned occupancy for subsequent reconstruction as $V(i,j,k)=\mathrm{Surf}(i,j,k)$.
\\

\textbf{LBVH construction.}
Following Karras~\cite{Karras2012Maximizing}, we build a GPU linear bounding volume hierarchy (LBVH) over per-Gaussian AABBs. Each Gaussian center $\mu_i$ is quantized inside the global scene AABB:
\begin{equation}
\begin{aligned}
\mathbf{u}_i &= \Big\lfloor 2^b (\mu_i-\mathbf{o})\oslash\mathbf{L}\Big\rfloor
\in \{0,\ldots,2^b-1\}^3, \\
m_i &= \mathrm{Morton}(\mathbf{u}_i).
\end{aligned}
\end{equation}
Here, $(\mathbf{o},\mathbf{L})$ denote the origin and size of the scene AABB, and $\oslash$ is elementwise division. We radix-sort primitives by $m_i$ and infer the hierarchy from the longest common prefix (LCP) of adjacent codes:
\begin{equation}
\mathrm{LCP}(i,j)=\max\{\ell:\text{ the first }\ell\text{ bits of }m_i, m_j\text{ match}\}.
\end{equation}
Node AABBs are then computed by parallel bottom-up reduction. The resulting hierarchy enables efficient Gaussian-to-occupancy conversion.

\textbf{Grid query and occupancy evaluation.}
The LBVH is used to accelerate density evaluation on the voxel grid by restricting each tile to a small set of candidate Gaussians. We partition the grid into tiles of size $B^3$ voxels, and for each tile AABB $\mathcal{B}_{\text{tile}}$ we cull candidate Gaussians by intersecting their AABBs.
\begin{equation}
\mathcal{C}_{\text{tile}}=\{\,G_i \mid b_i \cap \mathcal{B}_{\text{tile}}\neq\emptyset\,\}.
\end{equation}

This converts per-voxel accumulation from $O(N)$ to $O(|\mathcal{C}_{\text{tile}}|)$ with $|\mathcal{C}_{\text{tile}}|\!\ll\! N$ in practice, making dense voxelization feasible.

Each voxel \(v\in\mathcal{V}_{\text{tile}}\) accumulates density only over \(\mathcal{C}_{\text{tile}}\):
\begin{equation}
D(v)=\sum_{G_i\in\mathcal{C}_{\text{tile}}}
\exp\!\Big(-\tfrac12 (v-\mu_i)^\top \Sigma_i^{-1}(v-\mu_i)\Big)\,\alpha_i^2.
\end{equation}

\begin{figure}[!t]
  \centering
  \includegraphics[width=\linewidth]{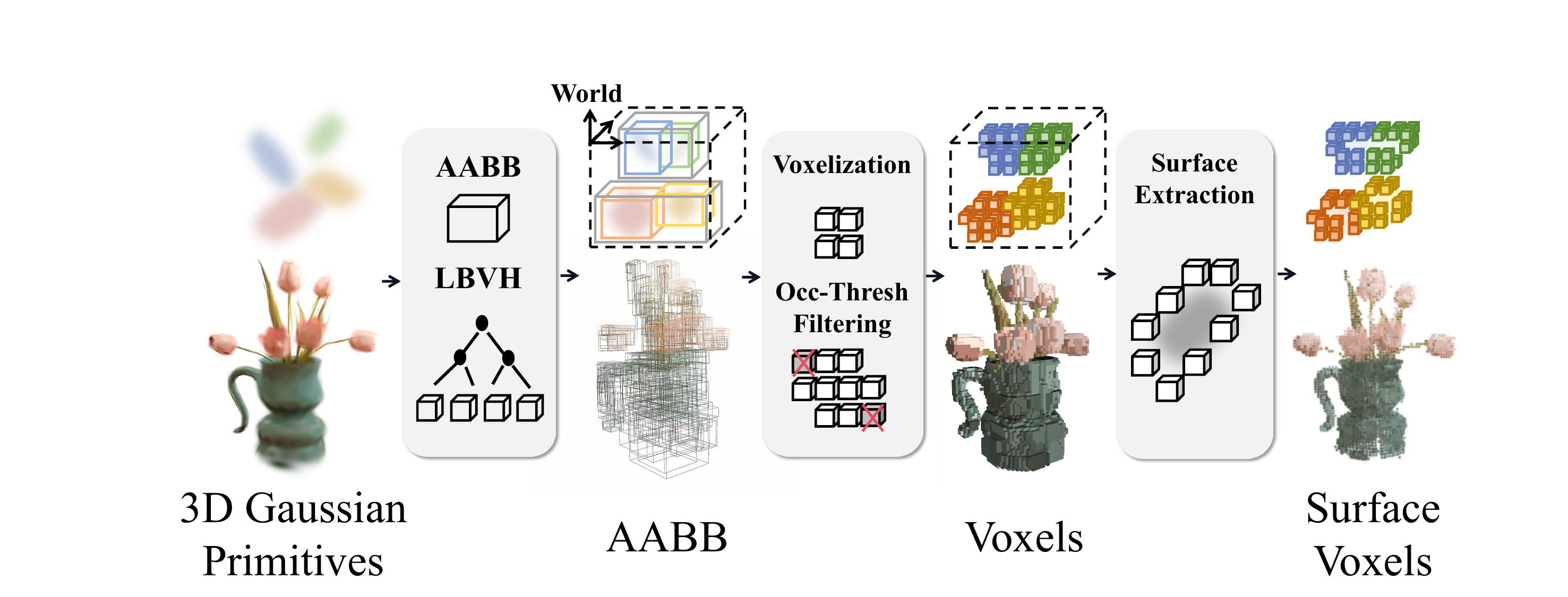}
    \caption{\textbf{LBVH-accelerated Gaussian-to-occupancy conversion.}
    Gaussian primitives are organized by an LBVH for efficient spatial querying, enabling localized voxel-density evaluation instead of global Gaussian--voxel accumulation. Thresholding and surface filtering then produce a surface-supporting occupancy representation for downstream reconstruction.}
  \label{fig:voxelization}
\end{figure}

We obtain a raw binary occupancy volume from $D(v)$ by thresholding. A voxel is marked occupied if
\begin{equation}
V_{\text{occ}}(i,j,k) =
\begin{cases}
1, & D(v_{ijk}) > \theta, \\
0, & \text{otherwise}.
\end{cases}
\end{equation}
where $\theta$ is a user-defined density threshold that controls the trade-off between completeness and sparsity.

We then remove interior voxels and retain only surface voxels to reduce reconstruction cost. A voxel is interior if its 6-neighborhood is fully occupied:
\begin{equation}
\Delta_6=\{(0,0,0),(\pm1,0,0),(0,\pm1,0),(0,0,\pm1)\}.
\end{equation}
\begin{equation}
\mathrm{Int}(i,j,k)=\bigwedge_{(a,b,c)\in\Delta_6} V_{\text{occ}}(i+a,j+b,k+c).
\end{equation}
The surface mask is then defined as
\begin{equation}
\mathrm{Surf}(i,j,k)=V_{\text{occ}}(i,j,k)\ \wedge\ \neg\,\mathrm{Int}(i,j,k),
\end{equation}
and we define the cleaned, surface-only occupancy used in Sec.~\ref{sec2} as $V(i,j,k)=\mathrm{Surf}(i,j,k)$. This surface-only occupancy reduces both memory and computation in the subsequent TSDF construction by restricting updates to a thin interface band.

\begin{figure}[h]
  \centering
  \includegraphics[width=\linewidth]{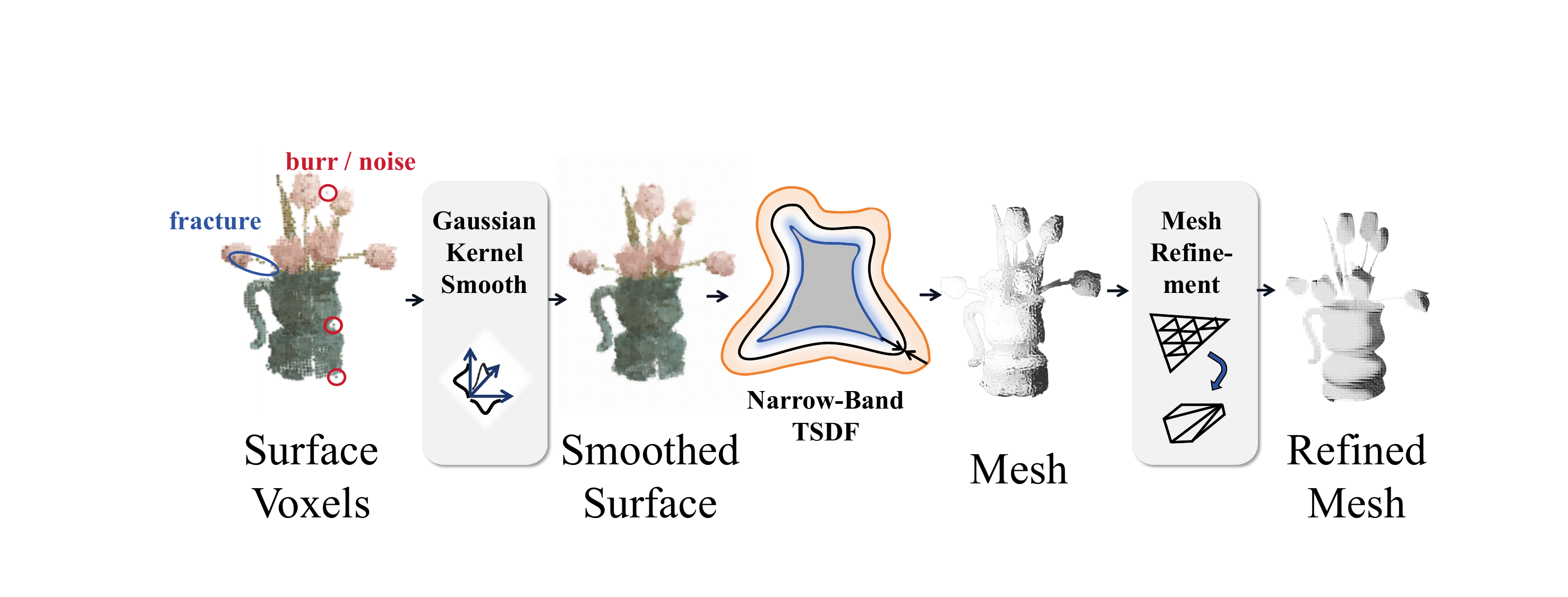}
\caption{\textbf{Simulation-oriented mesh reconstruction from surface occupancy.}
Surface-supporting voxels are denoised and converted into a narrow-band TSDF, from which a watertight isosurface is extracted. Mesh simplification and smoothing then reduce simulation cost while preserving the geometry needed for LiDAR first-hit queries.}

  \label{fig:denoised}
\end{figure}

\begin{table*}[t]
\vspace{2pt}
\centering
\caption{\textbf{LiDAR rendering latency (ms) under batched multi-environment settings, compared with Isaac Sim.}}
\label{tab:render_time_static}
\setlength{\tabcolsep}{4pt}
\small
\begin{tabular}{l cc cc cc cc cc}
\toprule
\multirow{2}{*}{Lines} & \multicolumn{10}{c}{Number of Environments} \\
\cmidrule(lr){2-11}
& \multicolumn{2}{c}{256} & \multicolumn{2}{c}{512} & \multicolumn{2}{c}{1024} & \multicolumn{2}{c}{2048} & \multicolumn{2}{c}{4096} \\
\cmidrule(lr){2-3}\cmidrule(lr){4-5}\cmidrule(lr){6-7}\cmidrule(lr){8-9}\cmidrule(lr){10-11}
& Ours & IsaacSim & Ours & IsaacSim & Ours & IsaacSim & Ours & IsaacSim & Ours & IsaacSim \\
\midrule
1k  & \textbf{0.9}$\pm$0.1  & 21$\pm$0.7
    & \textbf{1.0}$\pm$0.1  & 19$\pm$1.7
    & \textbf{2.3}$\pm$0.2  & 32$\pm$0.8
    & \textbf{4.6}$\pm$0.2  & 40$\pm$1.8
    & \textbf{5.0}$\pm$0.2  & 95$\pm$2.1 \\
4k  & \textbf{1.4}$\pm$0.2  & 28$\pm$0.6
    & \textbf{2.4}$\pm$0.1  & 39$\pm$0.3
    & \textbf{5.2}$\pm$0.1  & 81$\pm$1.1
    & \textbf{10.6}$\pm$0.2 & 146$\pm$4.8
    & \textbf{21.4}$\pm$0.5 & 308$\pm$5.8 \\
16k & \textbf{5.1}$\pm$0.3  & 81$\pm$3.1
    & \textbf{10.5}$\pm$0.3 & 142$\pm$1.4
    & \textbf{20.9}$\pm$0.5 & 317$\pm$17.3
    & \textbf{42.3}$\pm$0.8 & 600$\pm$8.8
    & \textbf{85.4}$\pm$1.6 & OOM \\
32k & \textbf{10.4}$\pm$0.2 & 148$\pm$2.3
    & \textbf{20.7}$\pm$0.2 & 286$\pm$2.7
    & \textbf{42.2}$\pm$0.7 & 569$\pm$14.9
    & \textbf{85.4}$\pm$1.3 & 1133$\pm$32.3
    & \textbf{170.8}$\pm$1.2 & OOM \\
\bottomrule
\end{tabular}
\end{table*}

\subsection{Mesh Reconstruction} \label{sec2}

The system requires a watertight surface for stable LiDAR first-hit queries, whereas the surface-only occupancy from Sec.~\ref{sec1} is still too discrete to serve directly as the simulation interface. We therefore reconstruct an implicit surface and extract a mesh representation suitable for downstream ray-casting.

Given the surface-only occupancy volume $V$ (after $\mathrm{Surf}$ filtering in Sec.~\ref{sec1}) with voxel spacing $\mathbf{s}=(s_x,s_y,s_z)$ (default $s_x=s_y=s_z=h$) and origin $\mathbf{o}$, we denoise and re-threshold the binary field, build a \emph{narrow-band} TSDF with reliable sign via outside flood-fill, extract an isosurface with Marching Cubes, and optimize the resulting mesh for both geometric fidelity and ray-casting efficiency.\\

\textbf{Binary denoising and re-thresholding.} To suppress salt-and-pepper artifacts while remaining resolution-agnostic, we convolve $V$ with a Gaussian kernel of metric scale $\sigma_{\text{smooth}}$ (expressed in meters and mapped to voxel units via $\mathbf{s}$):
\begin{equation}
V'(x) = (G_{\sigma_{\text{smooth}}} * V)(x).
\end{equation}
A denoised occupancy $\tilde V$ is obtained by either a fixed threshold $\tau$ or a quantile threshold $q$:
\begin{subequations}\label{eq:rethresh}
\begin{align}
\tilde V_{\text{fixed}}(x) &=
\begin{cases}
1, & V'(x) \ge \tau,\\
0, & \text{otherwise},
\end{cases}\\[4pt]
\tilde V_{\text{quant}}(x) &=
\begin{cases}
1, & V'(x) \ge \mathrm{Quantile}_q(V'),\\
0, & \text{otherwise}.
\end{cases}
\end{align}
\end{subequations} \\

\textbf{Narrow-band TSDF.}
A global distance transform is unnecessary for our goal, since only a thin band around the interface is needed for stable isosurface extraction. Reliable sign assignment is then critical for watertightness. Given a surface indicator grid $V:\Omega\to\{0,1\}$ where $V(x)=1$ denotes \emph{surface voxels} (Sec.~\ref{sec1}), we construct a signed distance field $\phi:\Omega\to[-r,r]$ in three logical stages (flooding in free space). We first assign signs by identifying the outside region: a flood-fill is performed over free voxels, seeded from a padded frame $\Gamma$ surrounding the domain, so that the 6-connected component $\mathcal{O}$ connected to $\Gamma$ is labeled as outside. Voxels in $\mathcal{O}$ are assigned $s(x)=+1$, while all others (occupied cells or enclosed voids) are assigned $s(x)=-1$, ensuring stable sign labeling even in the presence of cavities and tunnels. 

Next, unsigned distances are computed by layered propagation from the boundary set
\begin{equation}
\mathcal{S}_0=\{\,x\mid \exists\,y\in\mathcal{N}_6(x),\;V(x)\neq V(y)\,\},
\end{equation}
where each expansion shell $\mathcal{S}_m$ grows over the 6-neighborhood and newly visited voxels record their first-arrival shell index $p(x)$. The unsigned distance is then approximated as
\begin{equation}
\delta(x)=p(x)\,v_{\min}, \qquad v_{\min}=\min(s_x,s_y,s_z),
\end{equation}
which provides a conservative lower bound of the Euclidean distance and prevents diagonal leakage on anisotropic grids. 

Finally, the signed distance field is obtained by combining the seeded sign and unsigned distance with truncation,
\begin{equation}
\phi(x)=\mathrm{clip}\big(s(x)\,\delta(x),-r,r\big),
\end{equation}
which restricts values to the radius-$r$ band and avoids the memory and time overhead of a global Euclidean distance transform. All steps are executed in parallel on the GPU, avoiding the overhead of full-grid distance transforms.\\

\textbf{Isosurface extraction.} We extract a smooth surface from the grid by computing the level set
\begin{equation}
\mathcal{S}=\{\,x\in\Omega \mid \phi(x)=\mathrm{iso}\,\},
\end{equation}
with $\mathrm{iso}=0$ by default (optional millimeter-scale bias $\pm\mathrm{iso}$). 
We discretize $\mathcal{S}$ via Marching Cubes with step-size parameter $\texttt{mc\_step}$ (default $\texttt{mc\_step}=1$) and map vertices to world coordinates using $\mathbf{o}$ and $\mathbf{s}$. 
Per-vertex normals are estimated from $\nabla\phi$ for consistent outward orientation, yielding a watertight raw mesh $\mathcal{M}_\mathrm{raw}$. \\

\textbf{Mesh optimization.}
To make the reconstruction scalable for large scenes, we apply quadric-error decimation to obtain $\mathcal{M}_\mathrm{simp}$, targeting a prescribed face count or ratio while removing tiny components.
We then perform Taubin non-shrinking smoothing on $\mathcal{M}_\mathrm{simp}$ with parameters $(\lambda,\mu)$ and $n_{\text{smooth}}$ iterations, reducing staircase/normal noise while preserving sharp features and thin walls:
\[
\mathcal{M}_\mathrm{final} 
= \mathrm{Smooth}_{\lambda,\mu,n_{\text{smooth}}}\big(\mathrm{Simplify}(\mathcal{M}_\mathrm{raw})\big).
\]
Unless stated otherwise, all scale parameters $(\sigma_{\text{smooth}}, r, \mathrm{iso})$ are specified in meters through $\mathbf{s}$, decoupling control from voxel resolution. This reduces triangle count and directly improves the cost of the ray-casting stage in Sec.~\ref{sec_lidar}.\\

\begin{table*}[t]
\vspace{2pt}
\centering
\caption{\textbf{Ablations of system components (Block A) and post-processing (Block B).}
VoxPts in K, Faces in M, $T$ in seconds; Peak GPU Mem is maximum process GPU memory.
For `w/o surface extraction', we use full occupancy; `w/o narrow-band TSDF' denotes global TSDF.
Removing LBVH is infeasible end-to-end; a micro-benchmark shows $\sim 10^3\times$ speedup over brute-force accumulation on sampled queries.}
\label{tab:ablation}

\small
\setlength{\tabcolsep}{4pt}
\renewcommand{\arraystretch}{1.12}
\begin{tabular}{l|c|c|c|c|c|c}
\hline
Variant & VoxPts (K) & Faces (M) & $T$ (s) & Peak GPU Mem (GB) & C-D $\downarrow$ & F-score $\uparrow$ \\
\hline
\multicolumn{7}{c}{\textbf{Block A: System ablations; post-processing fixed}} \\
\hline
Full pipeline
& 8192.2 & 1.478 & 60 & 22 & \snd{0.006879} & \snd{0.9970} \\
w/o surface extraction
& 21274.7 & 1.462 & 90 & 26 & \best{0.006671} & \best{0.9975} \\
w/o narrow-band TSDF
& 8192.2 & 1.478 & $>600$ & 22 & \snd{0.006879} & \snd{0.9970} \\
w/o flood-based sign seeding
& 8192.2 & 1.964 & 80 & 20.65 & 0.013901 & 0.9943 \\
w/o LBVH
& -- & -- & -- & -- & -- & -- \\
\hline
\multicolumn{7}{c}{\textbf{Block B: Post-processing ablations; pipeline fixed}} \\
\hline
w/o decimation, w/o smoothing
& 8192.2 & 16.707 & 55 & 22 & 0.015538 & 0.9929 \\
w/o decimation, w/ smoothing
& 8192.2 & 16.761 & 80 & 22 & 0.018160 & 0.9922 \\
w/ decimation, w/o smoothing
& 8192.2 & 1.478 & 57 & 22 & 0.008200 & 0.9959 \\
\hline
\end{tabular}
\end{table*}

\subsection{Ray-casting for LiDAR Simulation}  \label{sec_lidar}
The system ultimately requires an efficient LiDAR simulation mechanism, since LiDAR sensing consists of a large number of repeated first-hit queries whose cost must remain low to approach real-time simulation. We therefore implement LiDAR simulation as GPU ray-casting and organize ray traversal with the same hierarchy-based acceleration principle as in Sec.~\ref{sec1}. Beyond efficient single-environment simulation, this design also supports batched multi-environment execution. \\
\textbf{LiDAR measurement model.} In a LiDAR scan, the $j$-th beam is modeled as a ray
\begin{equation}
r_j(t) = x_s + t\, d_j, \quad t \ge 0,
\end{equation}
where $T_s=\begin{bmatrix}R_s & t_s\\ \mathbf{0}^\top & 1\end{bmatrix}\in SE(3)$ is the sensor pose in world coordinates, $x_s:=t_s$ is the beam origin, and $d_j\in\mathbb{S}^2$ is a unit direction determined by the scanning pattern.  
Under this simulation interface, the environment is represented as a triangle mesh
\[
\mathcal{M}=\{\triangle_k=(v_{k,0},v_{k,1},v_{k,2})\}_{k=1}^{N_\triangle}.
\]
For each ray, the LiDAR return corresponds to the nearest intersection
\begin{equation}
\small
t_j^\star = \min \Big\{\, \eta(r_j,\triangle_k)\ \Big|\ 1\le k\le N_\triangle \Big\},
\end{equation}
where $\eta(r_j,\triangle_k)$ is the ray--triangle intersection parameter (discarding misses and intersections outside $[t_{\min},t_{\max}]$), and $t_{\min},t_{\max}$ are the sensor's minimum and maximum ranges. \\

\textbf{GPU-accelerated ray-casting.}
We implement ray-casting entirely on the GPU, assigning one thread to each LiDAR beam to exploit beam-level parallelism. In our design, every thread traverses a triangle-LBVH built over mesh triangle AABBs, constructed using the same GPU LBVH routine as Sec.~\ref{sec1} but with triangle primitives. Traversal is strictly guided by the best-so-far depth $t_j^\star$: nodes whose entry distance exceeds $t_j^\star$ are discarded together with their subtrees, and candidate triangles lying beyond this threshold are likewise excluded. This early-termination mechanism ensures that computation remains focused only on geometrically relevant regions. The kernels are designed for batched multi-environment execution, enabling high-throughput LiDAR simulation across thousands of parallel environments on a single GPU.

Formally, the per-ray work can be written as
\begin{equation}
\mathrm{cost}(r_j) = C_{\mathrm{trav}}\; N_{\mathrm{nodes}}(r_j) + C_{\mathrm{tri}}\; K_j,
\end{equation}
where $C_{\mathrm{trav}}$ and $C_{\mathrm{tri}}$ are the costs of a ray--AABB and ray--triangle test, $N_{\mathrm{nodes}}(r_j)$ is the number of BVH nodes visited, and $K_j$ is the number of triangles tested at leaves.  
Summing over all rays gives
\begin{equation}
\mathcal{C}_{\mathrm{GPU}} = \sum_{j=1}^{N_r} \mathrm{cost}(r_j) \;\approx\; \mathcal{O}(N_r\cdot(\log N_\triangle + \overline{K})),
\end{equation}
with $\overline{K}=\tfrac{1}{N_r}\sum_j K_j \ll N_\triangle$ in practice.  
This stands in contrast to the naive baseline
\[
\mathcal{C}_{\mathrm{naive}} = \mathcal{O}(N_r \cdot N_\triangle),
\]
showing that the GPU design substantially reduces per-scan cost even for million-triangle meshes.

Beyond hierarchical pruning, we further optimize memory layout and traversal coherence for GPU efficiency, reducing memory traffic and branch divergence in batched ray-casting.

\vspace{-2pt}

\section{experiments}

\subsection{Experiment Setup}

\textbf{Datasets.}
We evaluate LiDAR-simulation fidelity on two real-capture scenes: \emph{Indoor-RealLiDAR} and \emph{Outdoor-RealLiDAR}.
For each scene, we obtain a GT watertight mesh from real LiDAR scans (SLAM-based reconstruction) and use the corresponding pretrained 3DGS asset provided by the capture system.
We render one LiDAR frame per scene for three sensor configurations (HDL64/OS128/VLP32) under the same sensor pose and scan pattern.

\emph{For baseline comparison}, we additionally use two COLMAP-posed indoor scenes (\emph{Indoor-COLMAP-1/2}) to run pose-dependent baselines under their default protocol (COLMAP poses are not used by our method).

\textbf{Competitors.}
We compare our method with two categories of baselines.
(i) \textbf{GT mesh (oracle GT simulation):} we render a LiDAR frame from the watertight GT mesh reconstructed from real LiDAR scans, using the same sensor pose and scan pattern.
(ii) \textbf{3DGS-to-mesh baselines:} on \emph{Indoor-COLMAP-1/2}, we run representative 3DGS-to-mesh methods (GS2Mesh and MILo) to obtain meshes from the same 3DGS asset, and render LiDAR frames using identical extrinsics and beam layouts.
We use the authors' official implementations with their recommended default settings, fusing depth/occupancy rendered from 400 COLMAP-posed views.

\textbf{Metrics.}
We report symmetric Chamfer Distance (CD) and F-score at 0.05\,m, per LiDAR and averaged over HDL64/OS128/VLP32.

\textbf{Implementation.}
Intel W3545 (3.2\,GHz) + NVIDIA RTX 6000 Ada Generation GPU. Voxel size 0.015\,m; TSDF truncation 0.06\,m.
Our default post-processing applies decimation (face ratio 0.1, 3 rounds, boundary protected) followed by two-step non-shrinking Laplacian smoothing (Taubin-style) with $(\lambda,\mu)=(0.5,-0.53)$ (boundary preserved).
Mesh conversion time $T$ measures end-to-end 3DGS$\rightarrow$mesh (excluding LiDAR rendering); for pose-dependent baselines, $T$ includes view rendering and fusion (e.g., 400 views).

\begin{table}[t]
\vspace{2pt}
\centering
\small
\setlength{\tabcolsep}{5pt}
\caption{\textbf{LiDAR-simulation fidelity vs.\ GT mesh simulation.}
We report CD and F-score at 0.05\,m for Indoor-RealLiDAR and Outdoor-RealLiDAR.}
\label{tab:gt_fidelity}
\begin{tabular}{l|cc|cc}
\toprule
\textbf{LiDAR} &
\multicolumn{2}{c|}{\textbf{Indoor-RealLiDAR}} &
\multicolumn{2}{c}{\textbf{Outdoor-RealLiDAR}} \\
& \textbf{CD} $\downarrow$ & \textbf{F-score} $\uparrow$
& \textbf{CD} $\downarrow$ & \textbf{F-score} $\uparrow$ \\
\midrule
HDL64 & 0.0034 & 0.9950 & 0.0157 & 0.9816 \\
OS128 & 0.0034 & 0.9950 & 0.0104 & 0.9867 \\
VLP32 & 0.0053 & 0.9918 & 0.0250 & 0.9789 \\
\midrule
Avg.  & 0.0041 & 0.9939 & 0.0170 & 0.9824 \\
\bottomrule
\end{tabular}
\end{table}

\subsection{Ablation Study}
\label{sec:ablation}

Table~\ref{tab:ablation} reports ablations of key system components (Block A) and post-processing choices (Block B). In Block A, we keep post-processing fixed and toggle one system component at a time; in Block B, we keep the upstream pipeline fixed and ablate decimation and smoothing. Removing surface extraction increases voxel points by $\sim$2.6$\times$ (8.2M$\rightarrow$21.3M) and slows conversion (60s$\rightarrow$90s) with only marginal fidelity change, but substantially increases memory pressure. Replacing narrow-band TSDF with global TSDF preserves quality but increases runtime from 60s to $>600$s, confirming that narrow-band truncation is essential for practical throughput. Turning off flood-based sign seeding degrades fidelity (CD 0.0139) and inflates faces (1.96M), indicating that robust inside/outside initialization is necessary to avoid topological artifacts affecting first-hit LiDAR returns. In post-processing, the raw Marching Cubes mesh is overly dense (16.7M faces) and less accurate; decimation is the main factor reducing complexity (to 1.48M faces) and improving fidelity, whereas smoothing without decimation is expensive and ineffective on ultra-dense meshes. Finally, end-to-end removal of LBVH is computationally infeasible, and a micro-benchmark shows $\sim10^3\times$ speedup over brute-force accumulation.

\vspace{-2pt}

\subsection{Dataset Comparisons}
\label{sec:dataset_comparisons}

\textbf{Comparison with GT.}
Table~\ref{tab:gt_fidelity} reports LiDAR-simulation fidelity against GT simulation across three LiDAR configurations.
On Indoor-RealLiDAR, our simulation closely matches GT (Avg.\ CD 0.0041\,m, F-score 0.9939).
On Outdoor-RealLiDAR, errors increase but remain within centimeter scale (Avg.\ CD 0.0170\,m, F-score 0.9824).

\textbf{Comparison with 3DGS-to-mesh baselines.}
On COLMAP-posed indoor benchmarks, FGGS-LiDAR achieves the lowest LiDAR-simulation error and the shortest reconstruction time among the evaluated methods (Table~\ref{tab:indoor_lidar_vlp32_reformatted}). MILo exhibits substantially higher errors, which are consistent with severe surface artifacts on large planar regions (e.g., floors and walls) and spurious protrusions that distort first-hit returns; GS2Mesh remains competitive but lags behind our method under the same protocol.

\textbf{Comparison with LiDAR-supervised methods.}
We further compare against the LiDAR-supervised simulators GS-LiDAR~\cite{jiang2025gslidar} and LiDAR-RT~\cite{zhou2024lidarrt}, which model intensity and ray-drop and require multi-frame supervision. Following their protocol, we sample 64 poses per scene, hold out 4 interior frames for evaluation, and use the remaining 60 as supervision for the baselines; ground truth comes from ray-casting each scene's watertight mesh under the same poses. Being geometry-only and unsupervised, FGGS-LiDAR needs no training yet reaches comparable accuracy from a single 3DGS asset (Table~\ref{tab:supervised_comparison}), bounded only by the input 3DGS coverage: unobserved regions cannot be recovered.

\begin{table}[t]
\vspace{2pt}
\centering
\small
\setlength{\tabcolsep}{5pt}
\caption{\textbf{LiDAR-supervised comparison on \emph{Indoor-COLMAP-1/2} (HDL-64).}
Mean and std of CD, and mean F-score over 4 held-out frames.}
\label{tab:supervised_comparison}
\begin{tabular}{l|cc|c}
\toprule
\textbf{Method} & \textbf{CD} $\downarrow$ & \textbf{F-score} $\uparrow$ & \textbf{Sup.} \\
\midrule
\multicolumn{4}{l}{\textbf{Indoor-COLMAP-1}} \\
Ours        & 0.0066$\pm$0.0013 & 0.9981 & None \\
GS-LiDAR    & 0.0010$\pm$0.0005 & 0.9996 & LiDAR \\
LiDAR-RT    & 0.0119$\pm$0.0053 & 0.9855 & LiDAR \\
\midrule
\multicolumn{4}{l}{\textbf{Indoor-COLMAP-2}} \\
Ours        & 0.0213$\pm$0.0094 & 0.9920 & None \\
GS-LiDAR    & 0.0006$\pm$0.0004 & 0.9995 & LiDAR \\
LiDAR-RT    & 0.0127$\pm$0.0073 & 0.9710 & LiDAR \\
\bottomrule
\end{tabular}
\end{table}

\subsection{LiDAR Simulation Performance}

We evaluate efficiency from two perspectives: (i) single-environment runtime scaling with mesh complexity across six 3DGS scenes, and (ii) batched rendering under large numbers of parallel environments.

\begin{figure}[t]
    \centering
    \includegraphics[width=\linewidth]{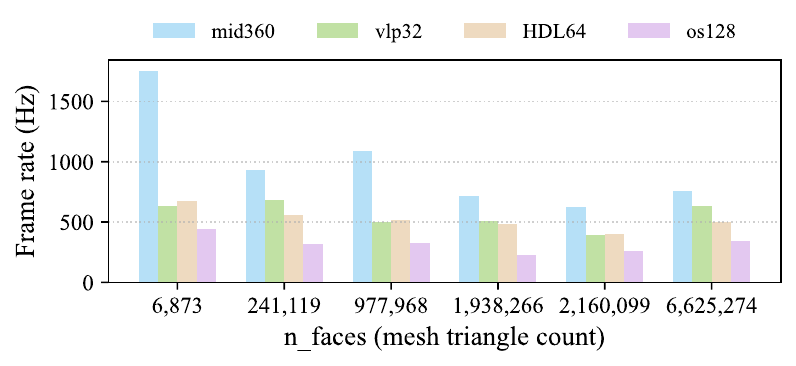}
    \caption{LiDAR frame rate vs.\ mesh complexity.}
    \label{fig:lidar_fps}
\end{figure}

\textbf{Frame rate.}
As shown in Fig.~\ref{fig:lidar_fps}, frame rate remains stable as mesh complexity increases to the multi-million-triangle regime, demonstrating well beyond real-time performance.

\begin{figure}[t]
    \centering
    \includegraphics[width=\linewidth]{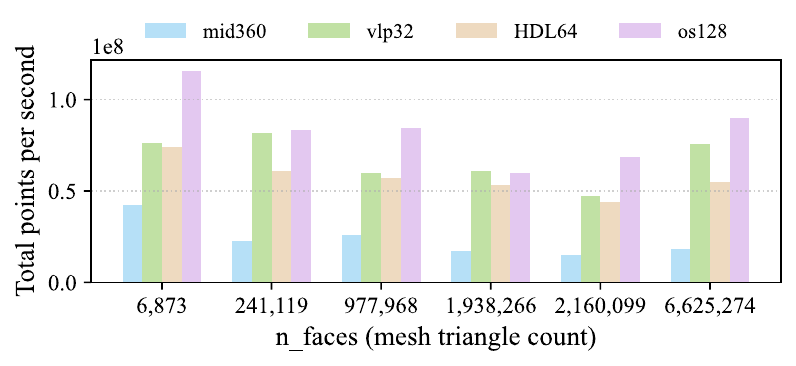}
    \caption{LiDAR throughput vs.\ mesh complexity.}
    \label{fig:lidar_throughput}
\end{figure}

\textbf{Throughput.}
Fig.~\ref{fig:lidar_throughput} reports point throughput versus mesh complexity. Denser sensors achieve higher throughput, exceeding $10^8$ points/s on lightweight meshes and sustaining $>7.5\times10^7$ points/s even at multi-million-triangle scales.

\textbf{Large-scale batched multi-environment performance.}
Table~\ref{tab:render_time_static} reports LiDAR rendering latency under increasing parallelism and beam counts, compared against Isaac Sim. Our simulator supports thousands of parallel environments on a single GPU and achieves order-of-magnitude lower latency (\(\sim 9\times\)–\(23\times\) speedup). Isaac Sim runs out of memory at the largest configurations, whereas our method remains feasible.

\begin{table}[t]
\vspace{4pt}
\centering
\caption{\textbf{Comparison with 3DGS-to-mesh baselines on indoor LiDAR simulation (VLP32).}
Results are on \emph{Indoor-COLMAP-1/2}; all methods use the same single-frame LiDAR pose, scan pattern, and beam layout. $T$ (min) is end-to-end 3DGS$\rightarrow$mesh time. Pose-dependent baselines follow the authors' default protocol and include view rendering and fusion (400 COLMAP-posed views).}
\label{tab:indoor_lidar_vlp32_reformatted}
\small
\setlength{\tabcolsep}{6pt}
\renewcommand{\arraystretch}{1.0}
\begin{tabular}{l | c c | c}
\toprule
\textbf{Method} & \textbf{CD} $\downarrow$ & \textbf{F-score} $\uparrow$ & \textbf{T (min)} $\downarrow$ \\
\midrule
\multicolumn{4}{l}{\textbf{Scene 1}} \\
Ours            & \best 0.0089  & \best 0.9957  & \best $<1$ \\
GS2Mesh         & 0.0126        & 0.9902        & 20 \\
MILo (std)      & 0.0523        & 0.8910        & 30 \\
MILo (highres)  & 0.1647        & 0.7625        & 120 \\
\midrule
\multicolumn{4}{l}{\textbf{Scene 2}} \\
Ours            & \best 0.0182  & \best 0.9815  & \best $<1$ \\
GS2Mesh         & 0.0816        & 0.9582        & 20 \\
MILo (std)      & 0.8206        & 0.8026        & 30 \\
MILo (highres)  & 0.3783        & 0.8042        & 120 \\
\bottomrule
\end{tabular}
\end{table}    

\vspace{-2pt}

\section{conclusion}

We present FGGS-LiDAR, an ultra-fast GPU-accelerated LiDAR simulation framework for general 3DGS assets. It operates directly on off-the-shelf 3DGS models without LiDAR supervision or external pose metadata by converting 3DGS into a watertight mesh via BVH-based volumetric discretization and narrow-band TSDF, followed by BVH-accelerated per-ray first-hit ranging. FGGS-LiDAR achieves over 500 FPS for 200k+ rays on a 6M+ triangle scene and provides 9--23$\times$ lower batched rendering latency than Isaac Sim up to 4096 environments. Experiments show strong agreement with GT simulation on indoor/outdoor scenes (4.1\,mm/17.0\,mm CD; 0.994/0.982 F-score) and competitive accuracy over prior 3DGS-to-mesh baselines on COLMAP-posed benchmarks. 

Remaining limitations include sensitivity to 3DGS quality and GPU memory usage; future work will improve scalability and incorporate richer LiDAR sensor physics.

\vspace{-1pt}

\balance
\bibliographystyle{ieeetr}
\bibliography{root}

\end{document}